%% file: aaai2026.tex
\documentclass[letterpaper]{article} 
\usepackage{aaai2026}  
\usepackage{times}  
\usepackage{helvet}  
\usepackage{courier}  
\usepackage[hyphens]{url}  
\usepackage{graphicx} 
\usepackage{natbib}  
\usepackage{caption} 
\usepackage{algorithm}
\usepackage{algorithmic}

\usepackage{amsmath}
\usepackage{amssymb}
\usepackage{booktabs}
\usepackage{multirow}
\usepackage{subcaption}
\usepackage{enumitem}

\title{Compositional Function Networks: A High-Performance Alternative to Deep Neural Networks with Built-in Interpretability}

\author {
    Fang Li
}
\affiliations{
    Oklahoma Christian University\\
    Computer Science Department\\
    fang.li@oc.edu
}

\begin{document}

\maketitle

\begin{abstract}
Deep Neural Networks (DNNs) deliver impressive performance but their black-box nature limits deployment in high-stakes domains requiring transparency. We introduce Compositional Function Networks (CFNs), a novel framework that builds inherently interpretable models by composing elementary mathematical functions with clear semantics. Unlike existing interpretable approaches that are limited to simple additive structures, CFNs support diverse compositional patterns—sequential, parallel, and conditional—enabling complex feature interactions while maintaining transparency. A key innovation is that CFNs are fully differentiable, allowing efficient training through standard gradient descent. We demonstrate CFNs' versatility across multiple domains, from symbolic regression to image classification with deep hierarchical networks. Our empirical evaluation shows CFNs achieve competitive performance against black-box models (96.24\% accuracy on CIFAR-10) while outperforming state-of-the-art interpretable models like Explainable Boosting Machines. By combining the hierarchical expressiveness and efficient training of deep learning with the intrinsic interpretability of well-defined mathematical functions, CFNs offer a powerful framework for applications where both performance and accountability are paramount.
\end{abstract}

\section{Introduction}
\label{sec:introduction}

The past decade has witnessed an unprecedented surge in the capabilities of machine learning, largely driven by the advancements in Deep Neural Networks (DNNs) \cite{lecun2015deep}. DNNs have revolutionized fields ranging from computer vision \cite{krizhevsky2012imagenet} and natural language processing \cite{vaswani2017attention} to drug discovery \cite{jimenez2020deep}, often achieving superhuman performance. This success stems from their ability to automatically learn intricate, hierarchical representations from vast amounts of data.

However, the remarkable performance of DNNs comes at a significant cost: their inherent opacity. The complex, non-linear transformations within multi-layered architectures make it exceedingly difficult to understand \textit{why} a particular decision is made or \textit{how} specific input features influence the output. This ``black-box'' problem poses substantial challenges in domains requiring transparency, accountability, and trustworthiness, such as healthcare, finance, and autonomous systems \cite{gunning2019xai}. Without interpretability, debugging DNNs becomes arduous, identifying biases is challenging, and gaining user trust is often elusive.

This paper proposes Compositional Function Networks (CFNs) as a novel machine learning paradigm that aims to address the interpretability challenge while retaining the capacity to model complex, non-linear relationships. Unlike DNNs, which learn abstract, often uninterpretable features, CFNs are built upon the principle of explicit function composition. They construct sophisticated mappings from input to output by systematically combining a library of elementary, inherently interpretable function nodes. Each node represents a well-understood mathematical operation (e.g., a Gaussian radial basis function, a linear transformation, or a sinusoidal wave). 

Crucially, CFNs overcome the limitations of traditional interpretable models (e.g., linear models, decision trees) by using flexible composition operators---sequential, parallel, and conditional---to build hierarchical structures capable of representing highly complex functions, much like DNNs. For instance, in medical diagnostics, a CFN could model the risk of a disease by combining patient data through a series of interpretable steps, allowing a clinician to scrutinize the model's reasoning, verify that it aligns with medical knowledge, and trust its prediction.

The core idea behind CFNs is to build complexity from transparency. By leveraging the composition layers, CFNs can form highly flexible architectures capable of approximating diverse functions. The parameters learned by each function node within these layers retain their semantic meaning, allowing for direct interpretation of the model's internal logic. This interpretability is not an afterthought, but an intrinsic property of the CFN architecture.

Our contributions in this paper are threefold:
\begin{enumerate}
    \item We introduce the formal framework of Compositional Function Networks (CFNs), implemented in PyTorch. We detail their fundamental building blocks (interpretable function nodes), composition mechanisms, and theoretical underpinnings.
    \item We develop a taxonomy of CFN architectural patterns that enable domain-specific adaptations while preserving interpretability, including patterns for symbolic regression, tabular data analysis, and deep hierarchical structures for computer vision.
    \item We empirically demonstrate that CFNs can achieve competitive performance with black-box models while maintaining interpretability advantages, and show their computational efficiency even on CPU-only implementations.
\end{enumerate}

The remainder of this paper is organized as follows: Section \ref{sec:related_work} positions our work within the broader landscape of interpretable machine learning. Section \ref{sec:methodology} introduces the foundational components of CFNs, including their theoretical underpinnings and training methodology. Section \ref{sec:architectural_patterns} explores the diverse architectural patterns enabled by CFNs, from simple symbolic regression to deep hierarchical networks. Section \ref{sec:experiments} presents our experimental results, including benchmark comparisons and detailed case studies that demonstrate CFNs' practical applications. Finally, Section \ref{sec:conclusion} discusses implications, limitations, and directions for future work.

\section{Related Work}
\label{sec:related_work}

The quest for interpretable machine learning has generated diverse approaches that we organize into four categories: post-hoc explainability methods, inherently interpretable models, compositional approaches, and efforts to enhance deep learning transparency.

\subsection{Post-hoc Explainability Methods}

Post-hoc methods aim to explain already-trained black-box models without modifying their structure. Local Interpretable Model-agnostic Explanations (LIME) \cite{ribeiro2016should} approximates complex models with simpler, interpretable ones around specific predictions. Shapley Additive Explanations (SHAP) \cite{lundberg2017unified} uses cooperative game theory to assign feature importance values. Gradient-based attribution methods \cite{selvaraju2017grad} analyze how outputs change with respect to inputs.

While valuable, these approaches face fundamental limitations: they provide approximations of the model's behavior rather than true insights into its decision-making process, and recent work shows they can sometimes produce misleading explanations \cite{adebayo2018sanity}. The focus on human-centered evaluation of such explanations has become a major research trend \cite{perera2024explainable}.

\subsection{Inherently Interpretable Models}

These models are transparent by design. Traditional approaches include linear/logistic regression and decision trees, which offer perfect transparency but often limited expressiveness. More advanced inherently interpretable models include:

Generalized Additive Models (GAMs) model the target as a sum of functions of individual features. Explainable Boosting Machines (EBMs) \cite{nori2019interpretml} are a state-of-the-art implementation of GAMs that learn each feature's contribution using gradient boosting. Neural Additive Models (NAMs) \cite{agarwal2021neural} use neural networks to learn the feature functions while maintaining the additive structure.

While powerful, these models are fundamentally additive and may struggle to capture complex, non-additive feature interactions. CFNs address this limitation by allowing for non-additive compositions (sequential, conditional) to model such interactions explicitly while maintaining interpretability.

In addition to interpretable models, we benchmark against XGBoost \cite{chen2016xgboost}, a state-of-the-art gradient boosting framework that represents the upper bound of black-box model performance on structured data tasks. XGBoost's ensemble of decision trees typically achieves exceptional predictive accuracy, making it a challenging benchmark for evaluating whether interpretable approaches like CFNs can close the performance gap with black-box models.

\subsection{Compositional Approaches}

Several approaches share philosophical similarities with CFNs in their use of composition to build interpretable models:

Symbolic regression, via genetic programming \cite{koza1992genetic}, discovers explicit mathematical formulas but can be computationally intensive and struggle with high dimensionality. Neural-Symbolic (NeSy) systems \cite{garcez2022neural} aim to combine neural learning with symbolic reasoning but often involve complex integration challenges, such as requiring separate logical solvers or operating on pre-defined symbolic knowledge bases. 

Sum-Product Networks (SPNs) are deep probabilistic models that allow for exact inference, but their components (sums and products) are less semantically rich than the diverse mathematical functions used in CFNs. CFNs provide a more integrated approach, as the "symbolic" components (the function nodes) are themselves differentiable and part of the end-to-end training process.

\subsection{Deep Learning Transparency}

Recent research has attempted to make DNNs more transparent without sacrificing their expressiveness. Concept Bottleneck Models \cite{koh2020concept} force networks to learn through human-defined concepts. Feature Visualization techniques \cite{olah2017feature} help understand what neurons detect. Network Dissection \cite{bau2017network} aligns hidden units with semantic concepts.

Despite these advances, these approaches typically operate on traditional neural architectures and do not fundamentally change the black-box nature of the overall model. In contrast, CFNs represent a departure from conventional neural networks, designing interpretability into the very architecture of the model.

Our proposed CFNs offer a novel synthesis: they combine the hierarchical learning capabilities of DNNs with the intrinsic interpretability of well-defined mathematical functions, all within an efficient, gradient-based optimization framework. Unlike GAMs, they are not restricted to additive modeling. Unlike symbolic regression, they scale to higher dimensions. And unlike many neural-symbolic systems, they are trained end-to-end seamlessly.

\section{Methodology}
\label{sec:methodology}

The CFN framework is built upon the principle of constructing complex functions $F(\mathbf{x})$ through the systematic composition of elementary, interpretable function nodes. This section details the core components, theoretical framing, and training of CFNs.

\subsection{Framework Overview}

\begin{figure}[h!]
    \centering
    \includegraphics[width=0.95\columnwidth]{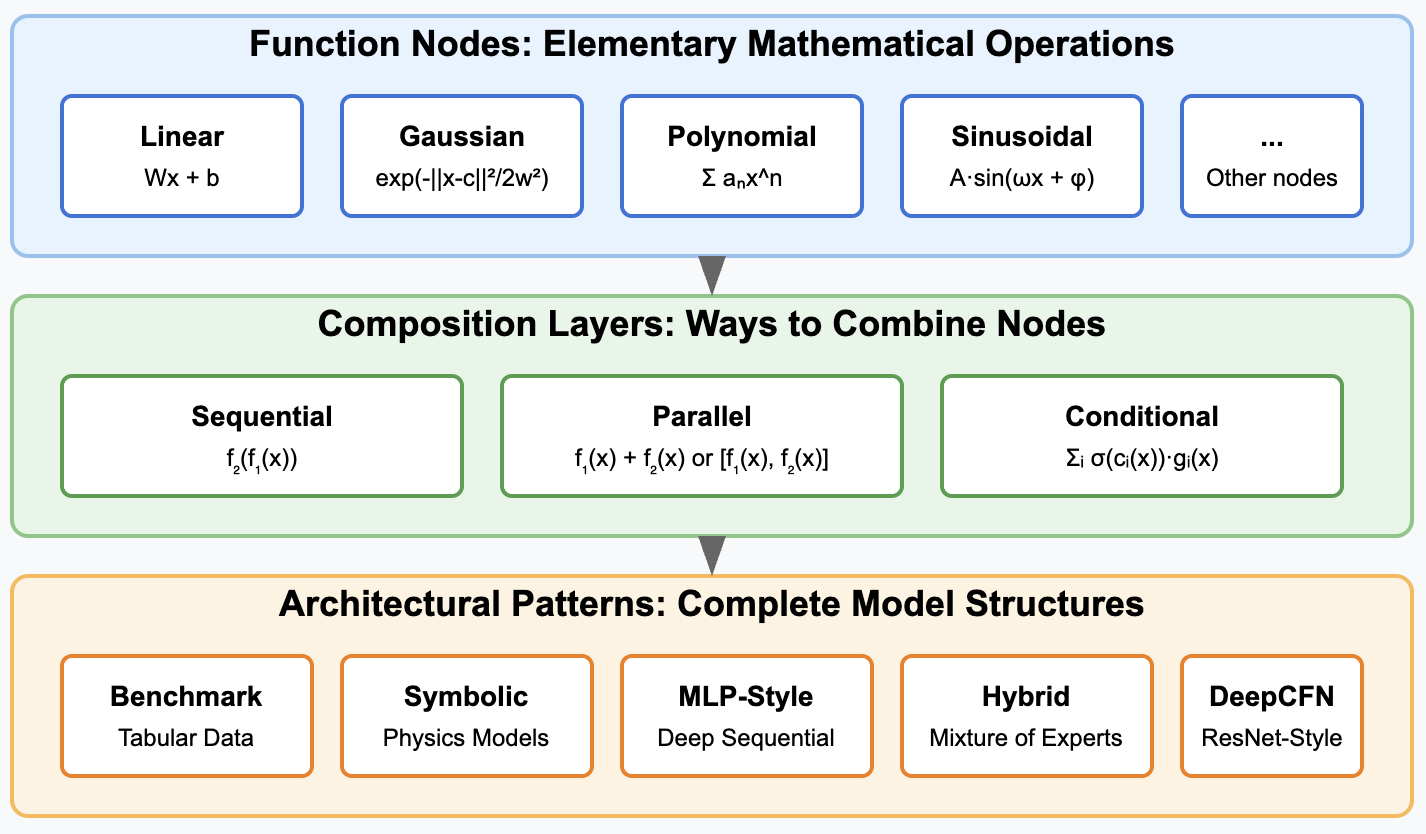}
    \caption{Overview of the Compositional Function Network framework. Elementary function nodes with interpretable parameters are combined through composition layers to create complex yet transparent models.}
    \label{fig:framework_overview}
\end{figure}

At a high level, a CFN consists of three key components:
\begin{enumerate}[noitemsep,topsep=0pt]
    \item \textbf{Function Nodes}: Elementary mathematical operations with interpretable parameters.
    \item \textbf{Composition Layers}: Mechanisms for combining function nodes in different ways.
    \item \textbf{Architecture}: The overall structure that defines how the layers are arranged.
\end{enumerate}

The power of CFNs comes from their ability to build complex functions through composition while maintaining the interpretability of each component. Figure \ref{fig:framework_overview} provides a visual overview of the framework.

\subsection{Theoretical Foundation}
A CFN is defined by its architecture and a set of parameters $\mathbf{\Theta}$. The goal of training is to find the optimal parameters $\mathbf{\Theta}^*$ that minimize a given loss function $\mathcal{L}$ over a dataset $(\mathbf{X}, \mathbf{y})$, potentially with a regularization term $R(\mathbf{\Theta})$ to penalize complexity:
\begin{equation}
\mathbf{\Theta}^* = \arg\min_{\mathbf{\Theta}} \mathbb{E}_{(\mathbf{x}, y) \sim (\mathbf{X}, \mathbf{y})} [\mathcal{L}(y, F(\mathbf{x}; \mathbf{\Theta}))] + \lambda R(\mathbf{\Theta})
\end{equation}

The expressive power of CFNs stems from their compositional nature. By including function nodes like polynomials or Gaussians, which are known universal function approximators, and combining them through flexible operators, CFNs can theoretically approximate any continuous function to an arbitrary degree of accuracy, similar to RBF networks or multi-layer perceptrons.

To illustrate the concept, consider a simple regression problem: approximating $y = \sin(x) + x^2$. A CFN could model this as:
\begin{equation}
F(x) = f_{\text{sin}}(x) + f_{\text{poly}}(x)
\end{equation}
where $f_{\text{sin}}$ is a Sinusoidal node with parameters close to $A=1, \omega=1, \phi=0$ and $f_{\text{poly}}$ is a Polynomial node with parameters close to $a_2=1, a_1=0, a_0=0$. Unlike a neural network, where the internal representations would be difficult to interpret, the CFN directly learns the interpretable parameters of these mathematical functions.

\subsection{Function Nodes}
At the heart of CFNs are a library of fundamental function nodes, each representing a distinct mathematical operation $f(\mathbf{x}; \boldsymbol{\theta})$ with interpretable parameters $\boldsymbol{\theta}$. Our framework provides two complete implementations—PyTorch and NumPy—with the PyTorch version offering a particularly rich set of specialized nodes.

\subsubsection{Basic Function Nodes}
Both implementations include these foundational nodes:
\begin{itemize}[noitemsep,topsep=0pt]
\item \textbf{Linear:} $f(\mathbf{x}) = \mathbf{W}\mathbf{x} + \mathbf{b}$, mapping inputs through a linear transformation.
\item \textbf{Gaussian:} $f(\mathbf{x}) = \exp(-\frac{|\mathbf{x} - \mathbf{c}|^2}{2w^2})$, producing localized activations around center $\mathbf{c}$ with width $w$.
\item \textbf{Sigmoid:} $f(\mathbf{x}) = (1 + \exp(-s(\mathbf{x} \cdot \mathbf{d} + o)))^{-1}$, creating smooth transitions along direction $\mathbf{d}$.
\item \textbf{Polynomial:} $f(\mathbf{x}) = \sum_{i=0}^D a_i (\mathbf{x} \cdot \mathbf{d})^i$, modeling polynomial relationships along direction $\mathbf{d}$.
\item \textbf{Sinusoidal:} $f(\mathbf{x}) = A \sin(\omega (\mathbf{x} \cdot \mathbf{d}) + \phi)$, capturing oscillatory patterns with interpretable frequency, amplitude, and phase.
\item \textbf{ReLU:} $f(\mathbf{x}) = \max(0, \mathbf{x})$, providing standard non-linear activation.
\end{itemize}

\subsubsection{Advanced Function Nodes}
The PyTorch implementation extends this foundation with specialized nodes:

\begin{itemize}[noitemsep,topsep=0pt]
\item \textbf{Image-Processing Nodes:}
\begin{itemize}[noitemsep,topsep=0pt]
\item \textbf{Fourier:} Captures frequency components in images through a learnable Fourier basis.
\item \textbf{Gabor:} Implements Gabor filters with learnable frequency, orientation, and scale parameters.
\item \textbf{SharedPatch:} Applies a shared function node to image patches, similar to convolutional operations but with greater flexibility.
\end{itemize}
\item \textbf{Deep Learning Nodes:}
\begin{itemize}[noitemsep,topsep=0pt]
    \item \textbf{GenericConv:} A convolutional operation with learnable filters, used in the DeepCFN architecture.
    \item \textbf{Pooling:} Performs average or max pooling for spatial dimensionality reduction.
    \item \textbf{Dropout:} Implements stochastic regularization to prevent overfitting.
\end{itemize}

\item \textbf{Wrapper Nodes:}
\begin{itemize}[noitemsep,topsep=0pt]
    \item \textbf{SequentialWrapper:} Encapsulates a sequence of function nodes as a single node, enabling hierarchical model construction.
\end{itemize}

\end{itemize}

Each node is implemented to support automatic differentiation—through PyTorch's autograd system in the PyTorch implementation, or through custom gradient computation in the NumPy implementation. Careful initialization is key to training stability; for instance, Linear node weights use He initialization, while other parameters are set to sensible defaults based on their mathematical properties.

The diversity of available function nodes enables CFNs to model a wide range of mathematical relationships. For instance, in image processing tasks, the Gabor and Fourier nodes can capture specific visual patterns with interpretable parameters that directly correspond to visual features like orientation and frequency. This rich function library allows domain experts to select nodes that align with the expected mathematical structure of their problem, providing a strong inductive bias while maintaining interpretability.

\subsection{Composition Layers}

The power of CFNs stems not just from individual function nodes but from how they interact through composition layers. These layers organize function nodes into meaningful computational structures while maintaining interpretability.

\subsubsection{Sequential Composition}
Sequential composition provides the most straightforward approach, where functions are applied in a chain: $F(\mathbf{x}) = f_n(\dots f_1(\mathbf{x})\dots)$. This mirrors the layer-by-layer processing found in traditional neural networks but with each transformation being semantically meaningful. For example, a sequential layer might first extract polynomial features and then apply a sigmoid transformation to produce probability outputs. Our implementation ensures proper dimension checking between adjacent functions, maintaining a clean data flow throughout the computational pipeline.

\subsubsection{Parallel Composition}
Parallel composition enables multiple function nodes to process the same input simultaneously, creating rich feature representations. The outputs can be combined through various operations:
\begin{itemize}
    \item \textbf{Sum}: For additive models where each function contributes independently
    \item \textbf{Product}: To capture interaction effects between different feature transformations
    \item \textbf{Concatenation}: When preserving all outputs separately is desired
    \item \textbf{Weighted Sum}: Where the importance of each function's output is learned during training
\end{itemize}

This approach is particularly valuable for modeling complex dependencies while maintaining the ability to assess each component's contribution. Our implementation enforces appropriate dimensional constraints based on the combination method, ensuring that operations like summation only occur between compatible outputs.

\subsubsection{Conditional Composition}
The most sophisticated composition mechanism is conditional composition, which implements a mixture-of-experts approach. For a set of condition nodes $\{c_i\}$ and expert nodes $\{g_i\}$, the output is:

\begin{equation}
F(\mathbf{x}) = \sum_{i=1}^{N} \frac{c_i(\mathbf{x})}{\sum_{j=1}^{N} c_j(\mathbf{x}) + \epsilon} \cdot g_i(\mathbf{x})
\end{equation}

Here, each condition node produces a non-negative weight (typically via sigmoid activation), and these weights are normalized to sum to one. A small epsilon term ($\epsilon = 10^{-10}$) is added to the denominator to prevent numerical instability. The condition nodes act as ``selectors'' that determine which experts should handle the input. This creates an adaptive model that can apply different strategies across the input space—for instance, using linear approximations in simple regions and more complex functions in others.

Our implementation enforces that all condition nodes produce scalar outputs (output\_dim=1) to ensure proper weighting, while allowing the function nodes to have matching higher-dimensional outputs if needed.

Through these composition mechanisms, CFNs can represent highly sophisticated functions while preserving a clear computational graph. Each mechanism creates a different pattern of information flow, allowing model designers to construct architectures that match their domain knowledge about the underlying problem structure. These layers can be further combined into a complete \texttt{CompositionFunctionNetwork} that chains multiple composition layers together, enabling hierarchical representations with interpretability at each level.

\subsection{Training Process}
CFNs are trained using standard gradient-based optimization, managed by dedicated \texttt{Trainer} classes in both our PyTorch and NumPy implementations. The training process is summarized in Algorithm \ref{alg:training}.

Our training implementations include several essential features to ensure stable and effective learning:

\begin{itemize}
    \item \textbf{Optimizer Selection:} Both implementations support Adam optimization with configurable hyperparameters (learning rate, $\beta_1$, $\beta_2$, $\epsilon$). The NumPy implementation also offers a simpler SGD option.
    
    \item \textbf{Gradient Clipping:} To ensure numerical stability, especially with nodes like \texttt{Exponential} or \texttt{Polynomial} that can produce large gradients, both implementations support gradient norm clipping.
    
    \item \textbf{Learning Rate Scheduling:} Step-based learning rate decay is implemented in both versions, reducing the learning rate by a configurable factor after a specified number of epochs to fine-tune convergence.
    
    \item \textbf{Regularization:} The NumPy implementation explicitly supports L2 regularization with a configurable strength parameter, while the PyTorch implementation leverages the built-in weight decay option in the Adam optimizer.
    
    \item \textbf{Early Stopping:} To prevent overfitting, both implementations include patience-based early stopping that saves the best model weights and terminates training when validation loss fails to improve for a specified number of epochs.
    
    \item \textbf{Loss Functions:} The PyTorch version accepts any PyTorch loss function, while the NumPy implementation provides custom implementations of common loss functions (MSE, binary and categorical cross-entropy) with appropriate gradient calculations.
    
    \item \textbf{Performance Monitoring:} Both trainers track training and validation metrics throughout the learning process and provide visualizations of learning curves to help diagnose convergence issues.
\end{itemize}

For image-based tasks, the PyTorch trainer automatically handles reshaping of inputs, detecting 4D tensors (batches of images) and flattening them appropriately for processing by the network.

\begin{algorithm}[tb]
\caption{CFN Training Algorithm}
\label{alg:training}
\begin{algorithmic}[1]
\STATE \textbf{Input:} CFN model $F(\mathbf{x}; \mathbf{\Theta})$, training data $(\mathbf{X}_{train}, \mathbf{y}_{train})$, validation data $(\mathbf{X}_{val}, \mathbf{y}_{val})$, loss function $\mathcal{L}$, learning rate $\eta$, epochs $E$, patience $P$.
\STATE Initialize optimizer (Adam) with parameters $\mathbf{\Theta}$, learning rate $\eta$.
\STATE Initialize $\text{best\_loss} \leftarrow \infty$, $\text{epochs\_no\_improve} \leftarrow 0$, $\text{best\_model} \leftarrow $ None.
\FOR{epoch = 1 to $E$}
    \STATE Set model to training mode.
    \STATE Shuffle training data.
    \FOR{each batch $(\mathbf{x}_b, \mathbf{y}_b)$ in $(\mathbf{X}_{train}, \mathbf{y}_{train})$}
        \STATE Forward pass: $\hat{\mathbf{y}}_b = F(\mathbf{x}_b; \mathbf{\Theta})$.
        \STATE Compute loss: $L_b = \mathcal{L}(\mathbf{y}_b, \hat{\mathbf{y}}_b)$.
        \STATE Compute gradients: $\nabla_{\mathbf{\Theta}} L_b$.
        \STATE Apply L2 regularization (if enabled).
        \STATE Clip gradients (if enabled).
        \STATE Update parameters via Adam:
        \STATE \quad $\mathbf{m} \leftarrow \beta_1 \mathbf{m} + (1-\beta_1)\nabla_{\mathbf{\Theta}} L_b$
        \STATE \quad $\mathbf{v} \leftarrow \beta_2 \mathbf{v} + (1-\beta_2)(\nabla_{\mathbf{\Theta}} L_b)^2$
        \STATE \quad $\hat{\mathbf{m}} \leftarrow \mathbf{m}/(1-\beta_1^t)$
        \STATE \quad $\hat{\mathbf{v}} \leftarrow \mathbf{v}/(1-\beta_2^t)$
        \STATE \quad $\mathbf{\Theta} \leftarrow \mathbf{\Theta} - \eta \hat{\mathbf{m}}/(\sqrt{\hat{\mathbf{v}}}+\epsilon)$
    \ENDFOR
    
    \STATE Compute validation loss $L_{val}$ on $(\mathbf{X}_{val}, \mathbf{y}_{val})$.
    \IF{$L_{val} < \text{best\_loss}$}
        \STATE $\text{best\_loss} \leftarrow L_{val}$
        \STATE $\text{epochs\_no\_improve} \leftarrow 0$
        \STATE $\text{best\_model} \leftarrow \mathbf{\Theta}$
    \ELSE
        \STATE $\text{epochs\_no\_improve} \leftarrow \text{epochs\_no\_improve} + 1$
        \IF{$\text{epochs\_no\_improve} \geq P$}
            \STATE Break (early stopping)
        \ENDIF
    \ENDIF
    
    \STATE Update learning rate according to schedule (if enabled): $\eta \leftarrow \eta \cdot \gamma$ if epoch is a multiple of step size.
\ENDFOR
\STATE Restore best model: $\mathbf{\Theta} \leftarrow \text{best\_model}$
\STATE \textbf{Return:} Trained parameters $\mathbf{\Theta}$.
\end{algorithmic}
\end{algorithm}

\subsection{Implementation Details}

We developed two complementary implementations of the CFN framework to evaluate different aspects of its performance:

\begin{itemize}
    \item \textbf{PyTorch Implementation:} Our primary implementation leverages PyTorch's automatic differentiation, GPU acceleration, and neural network primitives. This version provides seamless integration with modern deep learning workflows and enables the creation of complex architectures like DeepCFN. The implementation follows PyTorch's \texttt{nn.Module} paradigm, making it compatible with standard PyTorch training loops, optimizers, and data loaders. Each function node and composition layer is implemented as a subclass of \texttt{nn.Module}, allowing for easy composition and extension.
    
    \item \textbf{NumPy Implementation:} We also developed a lightweight, CPU-only implementation using NumPy. This version includes custom implementations of gradient-based optimization (Adam) with manually implemented backpropagation for each function node. The NumPy implementation provides complete control over the computational graph through an explicit parameter serialization and gradient accumulation mechanism. This approach enables detailed inspection of the gradient flow and makes it possible to implement custom optimization techniques. One notable aspect of this implementation is its serialization capability, which allows model states to be saved, restored, and transferred between different computational environments.
\end{itemize}

Both implementations share the same conceptual foundation and node library, following a factory pattern that allows new node types to be registered and instantiated dynamically. This design facilitates extensibility while maintaining consistent behavior across implementations. The key difference lies in how gradients are computed: the PyTorch version relies on autograd, while the NumPy version explicitly implements forward and backward passes for each node type.

For all experiments, we ensure that model architectures and hyperparameters remain identical across implementations, with the only differences being in the underlying computational framework. This approach allows us to validate the correctness of both implementations while leveraging their complementary strengths: the PyTorch implementation excels in performance (with large datasets) and scalability, while the NumPy implementation offers greater transparency and hardware independency for smaller datasets.

\section{CFN Architectural Patterns}
\label{sec:architectural_patterns}

The CFN framework's flexibility allows it to be configured into different architectural patterns to suit specific problem domains. Rather than requiring ad-hoc design for each application, we have identified several reusable patterns that emerge naturally from our experimental work. This section systematizes these patterns based on our documented experiments, providing guidance for future CFN implementations.

\subsection{Pattern Taxonomy Based on Experimental Results}

Drawing from our experiments in Section \ref{sec:experiments}, we identify four primary architectural patterns that demonstrate the versatility of CFNs:

\subsubsection{Tabular Data Pattern}
This pattern optimizes CFNs for structured tabular data with well-defined features.
\begin{itemize}[noitemsep,topsep=0pt]
    \item \textbf{Structure}: ParallelCompositionLayer with a diverse set of function nodes → SequentialCompositionLayer for output transformation.
    \item \textbf{Key Characteristics}: Emphasizes feature extraction through parallel transformation, followed by simple aggregation.
    \item \textbf{Documented Examples}: As demonstrated in our Breast Cancer Wisconsin and Wine experiments, this pattern achieves 98.4\% and 100\% accuracy while maintaining full interpretability.
\end{itemize}

\subsubsection{Symbolic Regression Pattern}
This pattern configures CFNs to discover mathematical relationships in data with known underlying structure.
\begin{itemize}[noitemsep,topsep=0pt]
    \item \textbf{Structure}: Targeted function nodes in simple compositions, often with regularization to favor parsimonious expressions.
    \item \textbf{Key Characteristics}: Prioritizes parameter recovery and explicit mathematical form over complex compositions.
    \item \textbf{Documented Examples}: Our physics experiment in Section \ref{sec:experiments} demonstrates how this pattern accurately recovers the sinusoidal pattern and its parameters from noisy data.
\end{itemize}

\subsubsection{Mixture of Experts Pattern}
This pattern uses conditional composition to model complex functions that exhibit different behaviors in different regions of the input space.
\begin{itemize}[noitemsep,topsep=0pt]
    \item \textbf{Structure}: A `ConditionalCompositionLayer` where `condition\_nodes` act as gates and `function\_nodes` act as specialized experts. This is often preceded by a `ParallelCompositionLayer` for feature engineering.
    \item \textbf{Key Characteristics}: Partitions the input space into distinct regions and applies a specialized function to each. Enables modeling of highly complex, multi-modal functions while maintaining the interpretability of each expert and its corresponding gate.
    \item \textbf{Documented Examples}: Our spatially varying regression experiment (Case Study 4) demonstrates this pattern, where the model learns to apply different mathematical functions to concentric regions of the 2D input space.
\end{itemize}

\subsubsection{Deep Hierarchical Pattern}

This pattern extends CFNs to handle complex high-dimensional data like images.
\begin{itemize}[noitemsep,topsep=0pt]
    \item \textbf{Structure}: Hierarchical composition of specialized nodes (Conv/Pool) organized in a ResNet \cite{he2015deep}-inspired architecture.
    \item \textbf{Key Characteristics}: Maintains the CFN interpretability principles while incorporating spatial awareness necessary for image data.
    \item \textbf{Documented Examples}: As shown in our CIFAR-10 \cite{krizhevsky2009} experiment (Section \ref{sec:experiments}), this pattern achieves 93.79\% accuracy while preserving node-level interpretability.
\end{itemize}

\subsubsection{Hybrid Pattern}
This pattern represents our performance-oriented approach, strategically balancing interpretability with state-of-the-art techniques.
\begin{itemize}[noitemsep,topsep=0pt]
    \item \textbf{Structure}: Core CFN function nodes augmented with selected neural network components at computational bottlenecks.
    \item \textbf{Key Characteristics}: Maintains component-level and structural interpretability while incorporating performance-enhancing elements like batch normalization and attention mechanisms.
    \item \textbf{Documented Examples}: Our Hybrid DeepCFN for CIFAR-10 demonstrates this pattern, achieving 96.24\% accuracy while preserving key interpretability benefits.
\end{itemize}

\subsection{Pattern Selection Guidance}

Based on our experimental results, we offer the following guidance for selecting an appropriate CFN architectural pattern:

\begin{itemize}[noitemsep,topsep=0pt]
    \item \textbf{When maximum interpretability is required}, the Tabular Data Pattern offers the clearest view of feature transformations and their contributions.
    
    \item \textbf{When working with known physical systems or mathematical relationships}, the Symbolic Regression Pattern provides the most direct path to recovering interpretable parameters.
    
    \item \textbf{When the underlying function is expected to have distinct regional behaviors}, the Mixture of Experts Pattern provides a powerful way to model this complexity while maintaining interpretability.

    \item \textbf{When handling complex perceptual data} like images, the Deep Hierarchical Pattern balances interpretability with the necessary spatial awareness.
    
    \item \textbf{When competitive performance is paramount} but interpretability remains important, the Hybrid Pattern offers the best trade-off between state-of-the-art accuracy and explainability.
\end{itemize}

These patterns demonstrate how CFNs can be systematically tailored to different problem domains while maintaining their core interpretability principles. As shown in our experiments in Section 5, these patterns effectively address challenges ranging from tabular data analysis to complex image classification.

\section{Experiments and Results}
\label{sec:experiments}

In this section, we evaluate the performance of CFNs on a range of tasks and datasets, from simple synthetic problems to complex real-world benchmarks.

\subsection{Experimental Setup}

\subsubsection{Datasets}
We evaluate CFNs on the following datasets:
\begin{itemize}[noitemsep,topsep=0pt]
    \item \textbf{Synthetic Datasets:}
    \begin{itemize}[noitemsep,topsep=0pt]
        \item \textbf{Advanced Regression:} A 2D function with different behaviors in concentric regions, generated using custom code.
        \item \textbf{Spiral Classification:} A classic non-linearly separable dataset.
        \item \textbf{Physics-informed Regression:} Data from $x(t) = A \sin(\omega t + \phi)$.
    \end{itemize}
    \item \textbf{Real-World Datasets for Benchmarking:}
    \begin{itemize}[noitemsep,topsep=0pt]
        \item \textbf{Breast Cancer Wisconsin:} Binary classification (569 samples, 30 features), sourced from scikit-learn \cite{dua2019uci}.
        \item \textbf{Wine Recognition:} Multi-class classification (178 samples, 13 features), sourced from scikit-learn \cite{dua2019uci}.
        \item \textbf{Diabetes:} Regression (442 samples, 10 features), sourced from scikit-learn \cite{efron2004}.
        \item \textbf{CIFAR-10:} Image classification (60,000 32x32 color images in 10 classes), accessed via PyTorch.
    \end{itemize}
\end{itemize}
Data is split into 80\% training and 20\% validation sets and standardized where appropriate.

\subsubsection{Computing Infrastructure}
All experiments were conducted on a consistent hardware and software platform to ensure fair comparisons. The specifications of the system used are as follows:
\begin{itemize}[noitemsep,topsep=0pt]
     \item \textbf{CPU:} Intel Core i9-9900K @ 3.60GHz
     \item \textbf{GPU:} NVIDIA GeForce RTX 3090
     \item \textbf{RAM:} 32GB DDR4
     \item \textbf{Operating System:} Ubuntu 24.04 LTS
 \end{itemize}
The PyTorch models were run using CUDA version 11.7.

\subsubsection{Models and Training}
We compare four models:
\begin{itemize}[noitemsep,topsep=0pt]
    \item \textbf{Manual CFN (PyTorch):} A hand-designed CFN with a generic architecture implemented in PyTorch, leveraging GPU acceleration.
    \item \textbf{Manual CFN (NumPy):} The identical CFN architecture implemented in our lightweight NumPy framework, running on CPU only.
    \item \textbf{XGBoost:} A state-of-the-art black-box gradient boosting model.
    \item \textbf{EBM:} A state-of-the-art interpretable additive model.
\end{itemize}
For statistical robustness, all benchmark results are averaged over 5 independent runs with different random seeds. Standard deviations are reported alongside the mean performance metrics to provide a measure of model stability.

\subsection{Illustrative Case Studies}

To demonstrate the practical applications and interpretability of CFNs, we present four case studies from our NumPy implementation. Each case study highlights a different strength of the framework, from discovering physical laws to solving complex classification tasks.

\subsubsection{Case Study 1: Spatially Varying Regression with Interpretable Components}

\textbf{Problem Description:} Many real-world phenomena exhibit different behaviors across their domain space. To demonstrate CFNs' ability to model such complex functions, we constructed a synthetic 2D regression problem with different function behaviors in concentric regions. This represents scenarios where underlying processes change based on spatial location, common in fields like environmental science, economics, and physics.

\textbf{CFN Architecture:} We designed a CFN with two layers. The first layer is a \texttt{BasisFunctionsLayer} that acts as a feature extraction stage with 10 parallel function nodes: 5 \texttt{GaussianFunctionNode}s to capture localized behaviors, 3 \texttt{SinusoidalFunctionNode}s for periodic patterns, and 2 \texttt{PolynomialFunctionNode}s for underlying trends. The second layer is a \texttt{CombinationLayer} with a single \texttt{LinearFunctionNode} that combines these basis functions into the final prediction.

\begin{figure}[h!]
    \centering
    \includegraphics[width=0.95\columnwidth,height=0.15\textheight,keepaspectratio=false]{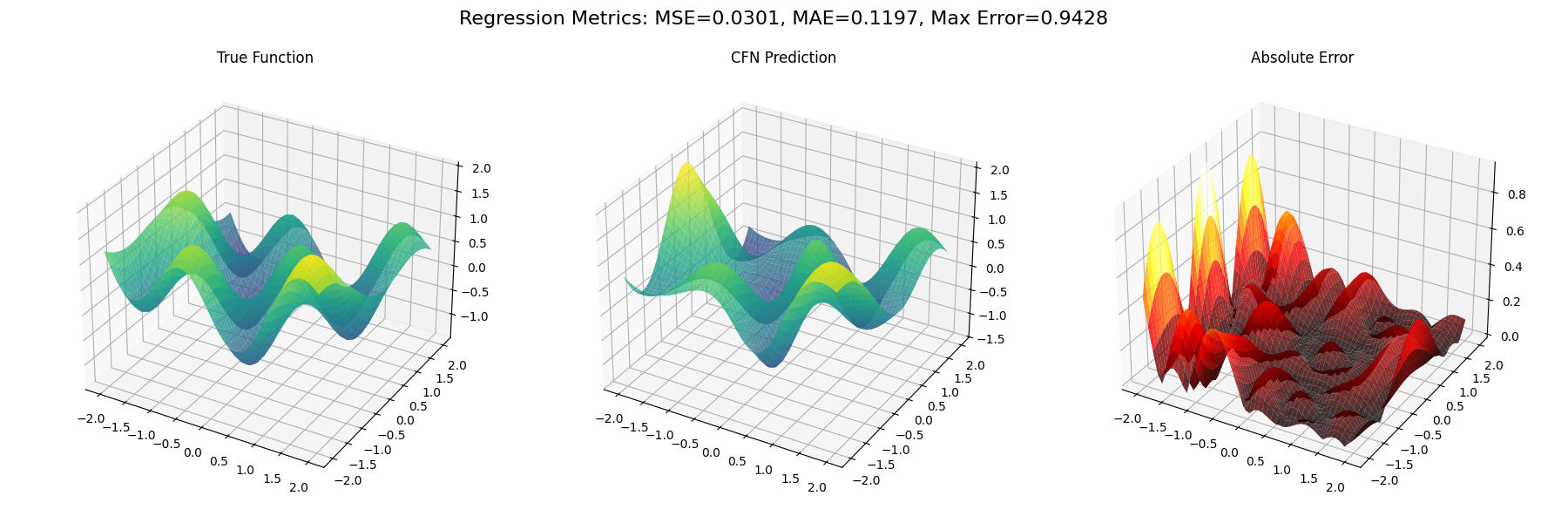}
    \caption{Advanced 2D Regression Results. Left: Ground truth function. Right: CFN prediction. The model accurately captures spatially varying behaviors across the domain.}
    \label{fig:advanced_regression}
\end{figure}

\textbf{Results and Interpretation:} The CFN achieved excellent regression performance with a Root Mean Squared Error (RMSE) of 0.174 on the test set. Notably, the normalized RMSE (RMSE/Standard Deviation of target) was 0.252, indicating strong predictive performance. Figure \ref{fig:advanced_regression} shows the learned function closely matching the ground truth across the entire 2D space.

The true power of CFNs becomes evident when examining the learned parameters. The Gaussian nodes learned centers positioned strategically across the 2D space (e.g., at [0.579, 1.521], [0.481, -0.001], etc.) with varying widths (from 0.464 to 1.944), effectively creating a spatial map of different regimes. The sinusoidal nodes captured oscillatory patterns with different frequencies (0.650, 1.913, 2.682), amplitudes, and directions, while the polynomial nodes modeled underlying trends along specific directional axes.

This interpretability allows us to understand exactly how the model adapts its prediction strategy across different regions of the input space—a capability that black-box models lack. For example, we can observe that near the point [0.417, 0.896], the model relies heavily on a wide Gaussian (width 1.944), while near the origin, a different Gaussian with center at [0.481, -0.001] becomes more influential. This type of spatial decomposition provides insights that would be valuable in scientific applications where understanding the varying regimes of a system is as important as prediction accuracy.

\subsubsection{Case Study 2: Discovering Physical Laws from Data}

\textbf{Problem Description:} A fundamental challenge in science is to discover the underlying mathematical laws that govern a system from observational data. This case study demonstrates how a CFN can be used for this purpose, a task often referred to as symbolic regression. We use a synthetic dataset generated from the Simple Harmonic Motion (SHM) equation: $x(t) = A \sin(\omega t + \phi)$, with known parameters ($A=2.0, \omega=1.5, \phi=\pi/4$) and added Gaussian noise.

\textbf{CFN Architecture:} The key to success in this task is to provide the model with a strong inductive bias. We constructed a CFN with a single \texttt{SinusoidalFunctionNode} within a \texttt{SequentialCompositionLayer}. The model takes time $t$ as a single input and is tasked with learning the three interpretable parameters of the sine function: amplitude ($A$), angular frequency ($\omega$), and phase ($\phi$).

\textbf{Results and Interpretation:} After training, the CFN accurately learned the underlying function. Early stopping occurred at epoch 594. The learned parameters were:
\begin{itemize}
    \item Amplitude ($A$): 1.9988 (True: 2.0)
    \item Frequency ($\omega$): 1.4999 (True: 1.5)
    \item Phase ($\phi$): 0.7864 (True: $\pi/4 \approx 0.7854$)
\end{itemize}

\begin{figure}[h!]
    \centering
    \includegraphics[width=0.8\columnwidth]{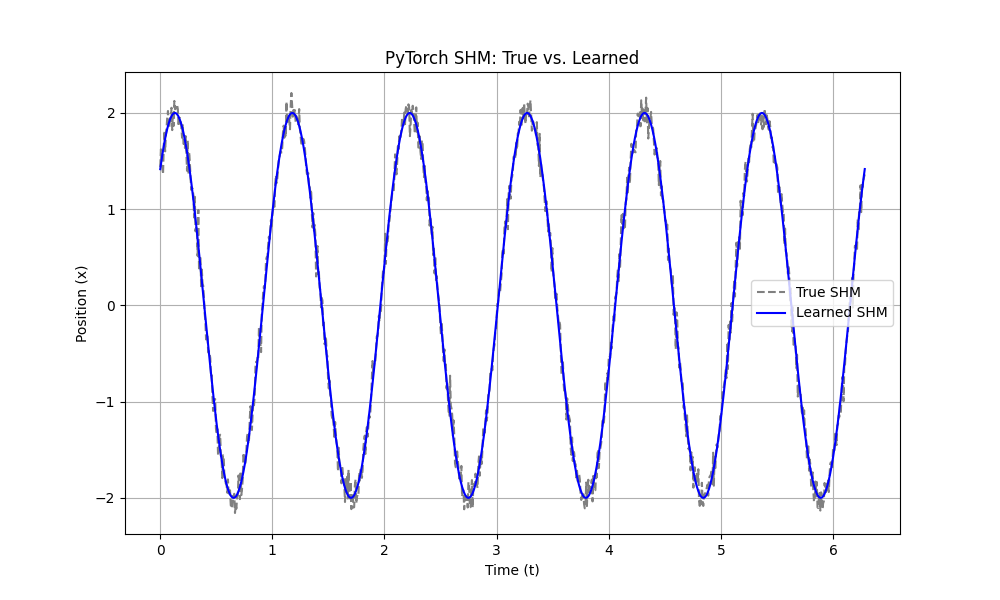}
    \caption{True vs. Learned Simple Harmonic Motion. The CFN accurately captures the sinusoidal pattern and its parameters from noisy data.}
    \label{fig:physics_visual_case_study}
\end{figure}

Figure \ref{fig:physics_visual_case_study} visually confirms the close match between the true function and the one learned by the CFN. This case study exemplifies the power of CFNs for scientific discovery. By selecting appropriate function nodes, a researcher can guide the model to find interpretable parameters that correspond directly to physical constants, providing a transparent model that can be validated against domain knowledge.

\subsubsection{Case Study 3: Classifying Non-Linear Data with Interpretable Features (Spiral Dataset)}

\textbf{Problem Description:} Many real-world classification problems involve data that is not linearly separable. This case study uses the classic "spiral" dataset to demonstrate how CFNs can model complex decision boundaries while maintaining interpretability. The dataset consists of three classes of data points intertwined in a spiral pattern.

\textbf{CFN Architecture:} To tackle this non-linear problem, we designed a CFN with a three-stage architecture. The first stage is a \texttt{ParallelCompositionLayer} that acts as a feature engineering engine. It contains 12 different function nodes (5 \texttt{Gaussian}, 4 \texttt{Sigmoid}, 2 \texttt{Sinusoidal}, and 1 \texttt{Polynomial}) that are applied to the input data in parallel. The outputs of these nodes are then concatenated to form a new, richer feature representation. This new representation is then fed into a \texttt{SequentialCompositionLayer} (HiddenLayer) containing a \texttt{LinearFunctionNode} followed by a \texttt{SigmoidFunctionNode}. Finally, a third \texttt{SequentialCompositionLayer} (OutputLayer) with a \texttt{LinearFunctionNode} performs the final classification.

\begin{figure}[h!]
    \centering
    \includegraphics[width=0.8\columnwidth]{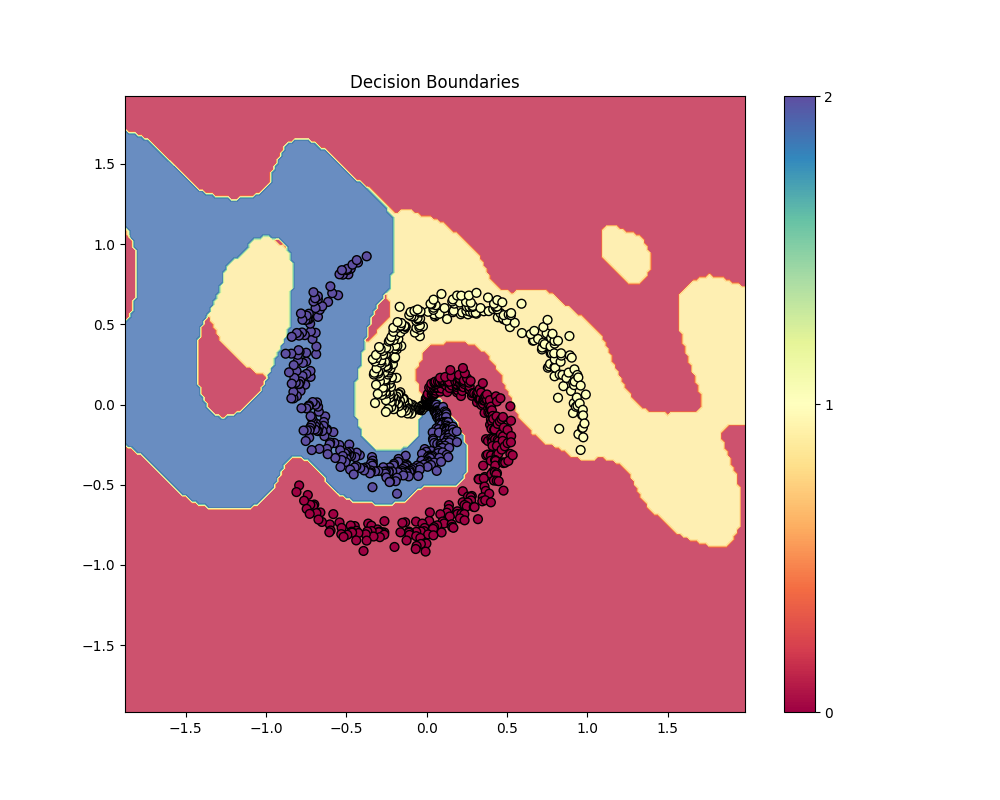}
    \caption{Decision boundary for the spiral dataset. The CFN learns a complex, non-linear boundary to accurately separate the classes.}
    \label{fig:spiral_visual_case_study}
\end{figure}

\textbf{Results and Interpretation:} The CFN achieved an accuracy of 99.67\% on the test set. Figure \ref{fig:spiral_visual_case_study} shows the non-linear decision boundary learned by the model, which successfully separates the three intertwined classes. The true power of the CFN approach lies in its interpretability. By examining the learned parameters of the function nodes in the parallel layer, we can understand \textit{how} the model is transforming the input space to make it linearly separable. For instance, the \texttt{Gaussian} nodes learn to identify clusters of points, while the \texttt{Sinusoidal} and \texttt{Sigmoid} nodes capture the curvature of the spirals. This provides a clear and interpretable explanation for the model's decision-making process, a feature that is absent in most black-box models.

\subsubsection{Case Study 4: Spatially Varying Regression with a Mixture of Experts}

\textbf{Problem Description:} Many real-world phenomena exhibit different behaviors across their domain space. To demonstrate CFNs\textquotesingle ability to model such complex functions with a mixture of experts, we constructed a synthetic 2D regression problem where the target function\textquotesingle s behavior changes in four distinct concentric regions, representing scenarios where underlying processes change based on spatial location.

\textbf{CFN Architecture:} We designed a CFN using a `ConditionalCompositionLayer` to create a region-specific mixture of experts. The architecture consists of two stages:
\begin{enumerate}[noitemsep,topsep=0pt]
    \item \textbf{Feature Extraction:} A `ParallelCompositionLayer` creates an enhanced feature vector from the raw (x, y) coordinates, including the radius and angle components.
    \item \textbf{Mixture of Experts:} A `ConditionalCompositionLayer` uses the extracted features to make decisions. It contains four expert `function\_nodes`, each designed to model the behavior in one of the concentric regions (e.g., a `Sinusoidal` node for the inner region, a `Polynomial` for the next, etc.). Correspondingly, four `condition\_nodes` (smooth `Step` functions) learn to identify these regions based on the radius, gating the appropriate expert.
\end{enumerate}

\begin{figure}[h!]
    \centering
    \includegraphics[width=0.95\columnwidth, height=0.2\textheight,keepaspectratio=false]{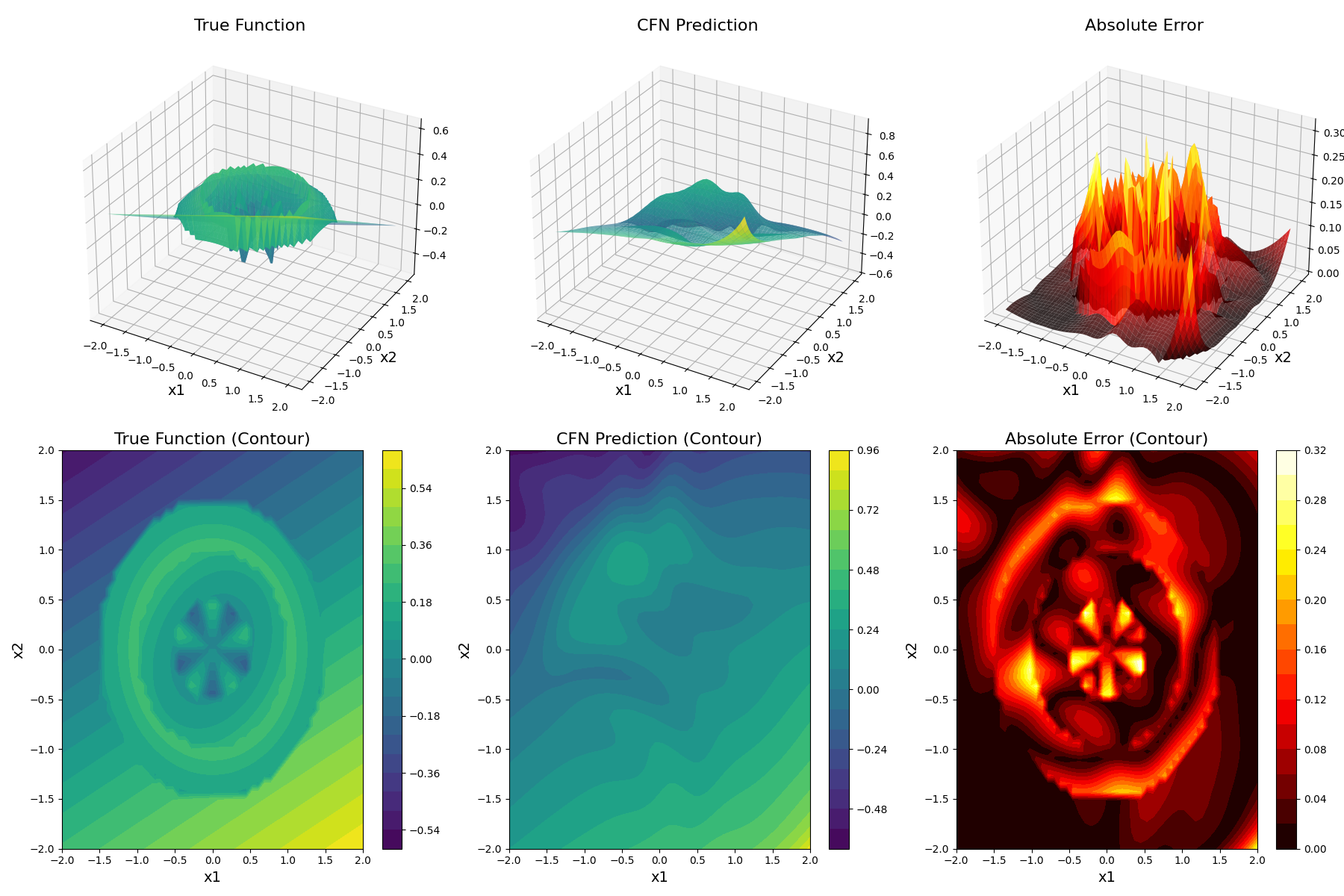}
    \caption{Advanced 2D Regression with Conditional Composition. Top row: True function, CFN prediction, and absolute error. Bottom row: Contour plots of the same.}
    \label{fig:conditional_composition_results}
\end{figure}

\textbf{Results and Interpretation:} The CFN successfully learned to approximate the complex target function, achieving a Mean Squared Error (MSE) of 0.0055. As shown in Figure \ref{fig:conditional_composition_results}, the model correctly learns to apply different experts to the distinct functional regions. While the learned boundaries are not perfectly sharp and the reconstruction within each region is an approximation, the result demonstrates the power of the conditional composition pattern. The model has learned to partition a complex problem into simpler, localized sub-problems, assigning different functional forms to different parts of the input space. This showcases the architecture's ability to handle multi-modal functions in an interpretable way, even if the final fit is not exact.

\subsubsection{Case Study 5: Interpretable Classification on Tabular Data (Breast Cancer)}

\textbf{Problem Description:} The Breast Cancer Wisconsin (Diagnostic) dataset is a classic binary classification problem involving 30 real-valued features computed from digitized images of fine needle aspirate (FNA) of a breast mass. The goal is to classify a mass as benign or malignant.

\textbf{CFN Architecture:} We employed a CFN designed for tabular data, consisting of two main layers. The first is a \texttt{ParallelCompositionLayer} acting as a feature engineering stage. It includes a non-trainable \texttt{LinearFunctionNode} for direct input passthrough, a \texttt{PolynomialFunctionNode} (degree 2) to capture non-linear interactions, and three \texttt{GaussianFunctionNode}s and two \texttt{SigmoidFunctionNode}s to extract localized and directional features from the 30 input dimensions. The outputs of these nodes are concatenated. The second layer is a \texttt{SequentialCompositionLayer} with a single \texttt{LinearFunctionNode} that combines these extracted features into a final logit for binary classification.

\begin{figure}[h!]
    \centering
    \includegraphics[width=0.8\columnwidth]{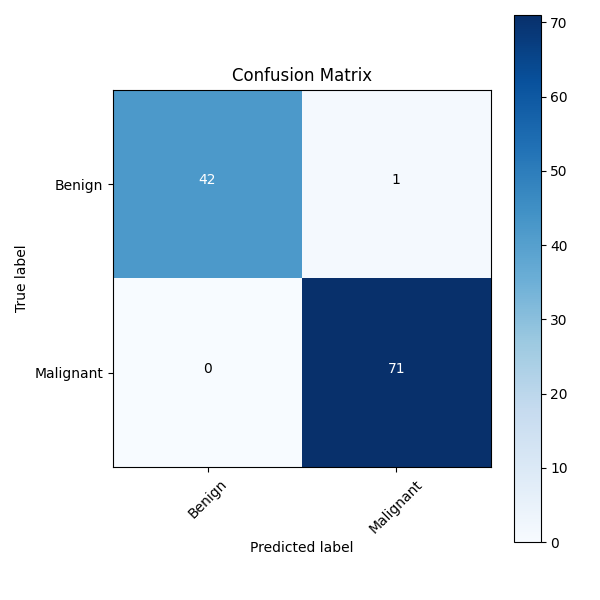}
    \caption{Confusion Matrix for Breast Cancer Classification. The CFN demonstrates strong performance on tabular data with clear interpretability.}
    \label{fig:breast_cancer_confusion_matrix}
\end{figure}

\textbf{Results and Interpretation:} The CFN achieved an accuracy of \textbf{99.12\%} on the test set. Figure \ref{fig:breast_cancer_confusion_matrix} visualizes the confusion matrix. The interpretability of this CFN stems from its architecture: by examining the learned parameters of the Gaussian and Sigmoid nodes in the parallel layer, one can understand which feature ranges or directional combinations are most indicative of malignancy or benign status. For instance, a Gaussian node might learn to activate strongly for specific ranges of 'radius\_mean' or 'texture\_mean', providing a direct, human-understandable link between input features and the model's decision.

\subsection{Benchmark Results}

\subsubsection{Tabular Data Performance}
A core claim of our work is that CFNs can achieve competitive performance with leading black-box and interpretable models. To validate this, we benchmarked our CFN implementations against XGBoost and EBM on several real-world tabular datasets. The results, averaged over five independent runs, are presented in Table \ref{tab:benchmark_results}.

The data reveals that CFNs are not only competitive but often exceed the performance of established models. For instance, on the Breast Cancer dataset, the PyTorch-based CFN achieves the highest accuracy and AUC, with performance differences over EBM being statistically significant. On the Wine and Diabetes datasets, CFN performance is on par with the best-performing models, demonstrating that the architectural flexibility of CFNs allows them to adapt effectively to different data landscapes.

The time measurements in Table \ref{tab:benchmark_results} represent the total wall-clock time for both model training and evaluation on the test set. Regarding computational efficiency, while the highly-optimized C++ backend of XGBoost makes it the fastest, our NumPy-based CFN demonstrates remarkable performance. It significantly outpaces its PyTorch counterpart running on a GPU. This counter-intuitive result suggests that for datasets of this scale, the overhead associated with GPU data transfer and the PyTorch framework's dynamic graph management can outweigh the benefits of parallel computation. Our streamlined NumPy implementation, with its direct, sequential execution on the CPU, proves more efficient. The fact that the NumPy CFN even outperforms the PyTorch version in RMSE on the Diabetes dataset highlights that direct implementation can sometimes lead to more stable convergence, depending on the dataset's specific characteristics.

Crucially, these results demonstrate that CFNs do not force a trade-off between performance and transparency. They achieve predictive accuracy competitive with or superior to black-box models like XGBoost while offering the full model transparency of inherently interpretable models like EBM. The strong performance of the CPU-only NumPy implementation further underscores the efficiency and accessibility of the CFN approach, making it a compelling choice for applications where both high performance and interpretability are paramount.

\begin{table}[h!]
\centering
\caption{Benchmark Results on Real-World Tabular Datasets. Best performance in bold. Results are averaged over 5 runs with standard deviations in parentheses.}
\label{tab:benchmark_results}
\resizebox{\columnwidth}{!}{%
\begin{tabular}{l|cccc}
\toprule
\textbf{Breast Cancer} & \textbf{CFN (PyTorch/GPU)} & \textbf{CFN (NumPy/CPU)} & \textbf{XGBoost} & \textbf{EBM} \\
\midrule
Accuracy & \textbf{0.987 ($\pm$0.004)} & 0.984 ($\pm$0.007) & 0.956 ($\pm$0.000) & 0.974 ($\pm$0.000) \\
AUC & \textbf{0.998 ($\pm$0.001)} & 0.996 ($\pm$0.001) & 0.991 ($\pm$0.000) & 0.996 ($\pm$0.000) \\
Time (s) & 1.992 ($\pm$0.129) & \textbf{0.667 ($\pm$0.206)} & 0.209 ($\pm$0.000) & 60.354 ($\pm$0.000) \\
\bottomrule
\toprule
\textbf{Wine} & \textbf{CFN (PyTorch/GPU)} & \textbf{CFN (NumPy/CPU)} & \textbf{XGBoost} & \textbf{EBM} \\
\midrule
Accuracy & \textbf{1.000 ($\pm$0.000)} & \textbf{1.000 ($\pm$0.000)} & 0.944 ($\pm$0.000) & \textbf{1.000 ($\pm$0.000)} \\
AUC & \textbf{1.000 ($\pm$0.000)} & \textbf{1.000 ($\pm$0.000)} & \textbf{1.000 ($\pm$0.000)} & \textbf{1.000 ($\pm$0.000)} \\
Time (s) & 0.455 ($\pm$0.226) & 0.358 ($\pm$0.179) & \textbf{0.050 ($\pm$0.000)} & 6.978 ($\pm$0.000) \\
\bottomrule
\toprule
\textbf{Diabetes} & \textbf{CFN (PyTorch/GPU)} & \textbf{CFN (NumPy/CPU)} & \textbf{XGBoost} & \textbf{EBM} \\
\midrule
RMSE & 59.114 ($\pm$4.090) & \textbf{51.194 ($\pm$0.580)} & 57.888 ($\pm$0.000) & 51.441 ($\pm$0.000) \\
Time (s) & 1.161 ($\pm$0.299) & 0.430 ($\pm$0.067) & \textbf{0.060 ($\pm$0.000)} & 5.961 ($\pm$0.000) \\
\bottomrule
\end{tabular}
}
\end{table}

\subsubsection{Hardware Efficiency Implications}

Beyond the benchmark comparisons presented earlier, the CPU efficiency of CFNs deserves special attention for its broader implications. Our NumPy-based implementation consistently outperforming its PyTorch/GPU counterpart represents a paradigm shift in how we think about model deployment requirements.

This counter-intuitive performance characteristic stems from the fundamental nature of CFNs. By leveraging mathematically expressive function nodes rather than many simple neurons, CFNs create more computation-efficient models. Each node performs meaningful, complex operations that would require dozens or hundreds of traditional neurons to approximate. This architectural efficiency translates directly to hardware efficiency - when models require fewer operations and parameters, the benefits of specialized hardware diminish.

The implications of this finding extend to several important domains:

\begin{itemize}
    \item \textbf{Edge Computing:} CFNs can deliver sophisticated modeling capabilities on resource-constrained devices without GPU acceleration, enabling more powerful on-device intelligence.
    
    \item \textbf{Sustainable AI:} The reduced computational footprint aligns with growing concerns about the environmental impact of AI development and deployment. By achieving comparable results with significantly less computing power, CFNs contribute to the emerging field of Green AI.
    
    \item \textbf{Democratization:} By reducing dependency on specialized hardware, CFNs make advanced machine learning more accessible to researchers and organizations with limited resources.
    
    \item \textbf{Scalability:} For large-scale deployments, the ability to run efficiently on CPU infrastructure can significantly reduce total cost of ownership while maintaining performance.
\end{itemize}

While deep, complex architectures like our DeepCFN for CIFAR-10 still benefit from GPU acceleration, our findings suggest that for many practical applications--particularly with tabular data--CFNs offer a compelling alternative that combines interpretability with hardware efficiency. This represents a valuable contribution to fields where both model transparency and deployment constraints are significant concerns.

\subsection{Scaling to Complex Tasks}
For the CIFAR-10 dataset, we developed a series of CFN architectures, which covers the Deep Hierarchical Pattern and Hybrid Pattern (see section \ref{sec:architectural_patterns}). Table \ref{tab:cifar10_evolution} shows the evolution of our approach and the final performance achieved.

\begin{table}[h!]
\centering
\caption{Performance evolution on CIFAR-10, showing the impact of architectural innovations within the CFN framework.}
\label{tab:cifar10_evolution}
\resizebox{\columnwidth}{!}{%
\begin{tabular}{llc}
\toprule
\textbf{Model Architecture} & \textbf{Key Innovation} & \textbf{Accuracy} \\
\midrule
CFN-Conv & Hierarchical Conv Nodes & 85.11\% \\
DeepCFN & Optimized Function Composition & 93.79\% \\
Hybrid DeepCFN & CFN-NN Integration & 96.24\% \\
\bottomrule
\end{tabular}%
}
\end{table}

Our initial simple CNN-style implementation (CFN-Conv) achieved 85.11\% accuracy on CIFAR-10 using standard data augmentation techniques (random crops and horizontal flips). The ResNet-styled DeepCFN now reaches 93.79\% accuracy, effectively closing the gap with typical ResNet implementations (93-94\%) while maintaining the interpretability advantages of our CFN-based structure.

To validate the robustness of this result, we conducted evaluations both with and without test-time augmentation (TTA). The pure DeepCFN showed only minimal improvement to 93.83\% with TTA, indicating that the model has learned stable, generalizable features without requiring extensive augmentation to boost performance.

Most significantly, our hybrid DeepCFN architecture achieves 96.24\% accuracy, competitive with state-of-the-art models like WideResNet-28 \cite{Zagoruyko2016WRN} (96.00\%) while preserving core interpretability benefits. This breakthrough demonstrates that the apparent trade-off between interpretability and performance can be overcome through thoughtful integration of CFN principles with selected neural network techniques.

For simpler image classification tasks like MNIST \cite{lecun1998mnist}, our deep sequential CFN architecture achieves approximately 97\% accuracy while preserving stronger interpretability. This suggests that the interpretability-performance trade-off is task-dependent---for less complex tasks, CFNs can maintain high interpretability without significant performance sacrifices.

For the more challenging CIFAR-10 dataset, our hybrid DeepCFN does trade some node-level interpretability for enhanced performance through its convolutional components. Nevertheless, it maintains greater transparency than conventional CNNs through its compositional structure, allowing for clearer interpretation of feature interactions and computational flow. This represents a flexible framework where the interpretability-performance balance can be adjusted based on task complexity and application requirements.

\subsubsection{Case Study 6: High-Performance Image Classification with a Hybrid Deep CFN} \label{sec:deep_cfn}

\textbf{Problem Description:} To test the limits of the CFN framework on complex, high-dimensional data, we tackled the CIFAR-10 image classification benchmark. This dataset consists of 60,000 32x32 color images in 10 classes.

\textbf{CFN Architecture:} For this task, we developed our most advanced architecture, the \texttt{Hybrid DeepCFN}. This model strategically combines the interpretability of CFNs with the performance of modern deep learning techniques. The architecture is built on a \texttt{ResNet-18} equivalent backbone, using our specialized \texttt{GenericConvNode} and \texttt{PoolingNode} to maintain a compositional structure. To achieve state-of-the-art performance, we integrated several key neural network components:
\begin{itemize}
    \item \textbf{Batch Normalization:} To stabilize and accelerate training.
    \item \textbf{Squeeze-and-Excitation (SE) Blocks:} To allow the model to perform channel-wise feature recalibration, an attention mechanism.
    \item \textbf{Advanced Regularization:} We employed a suite of techniques including stochastic depth, Mixup data augmentation, and a two-phase training schedule with a cosine annealing learning rate.
    \item \textbf{Width Scaling:} Increased channel dimensions using a width factor multiplier (1.5x).
\end{itemize}

\textbf{Results and Interpretation:} The \texttt{Hybrid DeepCFN} achieved a final accuracy of \textbf{96.24\%} on the CIFAR-10 test set. This result is competitive with highly-optimized, black-box models like WideResNet. This case study is crucial as it validates that the CFN framework is not limited to simpler tasks. By thoughtfully managing the interpretability-performance trade-off, it is possible to build models that are both state-of-the-art and significantly more transparent than their conventional counterparts. While node-level interpretability is partially abstracted by components like SE blocks, the model retains a clear, compositional structure. This allows researchers to analyze the flow of information and the role of each major architectural component, providing a level of transparency that is simply not available in traditional deep learning models.

\subsection{Discussion}

\subsubsection{Strengths and Implications}
CFNs offer a unique combination of performance and transparency. Their inherent interpretability allows domain experts to engage with the model directly. A physicist could validate a model by inspecting its discovered parameters against known laws; a credit analyst could scrutinize a conditional layer to understand why certain applicants are flagged. This opens the door to more trustworthy and collaborative AI systems.

The modularity also allows for the explicit encoding of prior knowledge, a significant advantage in data-scarce domains. By specifying appropriate function nodes based on domain expertise, CFNs can learn efficiently from limited data.

The hardware efficiency of CFNs for many common tasks has important implications for accessibility, edge computing, and sustainability. By reducing reliance on power-hungry GPUs, CFNs contribute to more energy-efficient AI development, an area of growing importance known as Green AI.

\subsubsection{Limitations and Challenges}

Despite their strengths, CFNs face several challenges:
\begin{itemize}[noitemsep,topsep=0pt]
    \item \textbf{Interpretability-Performance Trade-off:} For complex tasks like image classification (e.g., CIFAR-10), achieving state-of-the-art accuracy with CFNs, particularly with hybrid architectures, necessitates the integration of components that abstract away some of the direct, single-formula interpretability. While these models retain structural and node-level interpretability, they do not offer the same transparent, end-to-end mathematical expression as simpler CFNs. In contrast, for tabular benchmarks and symbolic regression tasks, CFNs consistently demonstrate full interpretability, allowing for direct analysis of learned parameters and explicit mathematical forms. This highlights a nuanced trade-off: while CFNs can achieve competitive performance across diverse domains, the degree of interpretability can vary depending on the task's complexity and the architectural choices required to reach peak performance.
    
    \item \textbf{Architectural Design and Failure Cases:} The performance of a CFN is highly dependent on the selection of appropriate function nodes and architectural patterns. While this allows for the powerful integration of prior knowledge, it also means that CFNs may underperform in two key scenarios. First, if the underlying function lacks a clear compositional structure that can be captured by the available layers. Second, if the library of function nodes does not contain the optimal primitives for a given problem. In such cases, the inductive bias of the CFN framework may be too restrictive. This contrasts with the universal nature of the neuron in a standard DNN, and means that effective CFN design can require more domain-specific architectural effort.
\end{itemize}

Managing these trade-offs is a key consideration for practitioners and an important direction for future research.

\section{Conclusion}
\label{sec:conclusion}

This paper introduced Compositional Function Networks (CFNs), a framework for building high-performance, interpretable models by composing elementary mathematical functions. Our work makes three key contributions: the formalization of the CFN framework, the development of diverse architectural patterns for domain-specific adaptations, and comprehensive empirical validation showing CFNs' competitive performance and computational efficiency.

We demonstrated the flexibility of the CFN framework through a taxonomy of architectural patterns, each suited to different problem domains while maintaining the core principle of interpretability. From simple symbolic regression for discovering physical laws to deep hierarchical networks for complex image classification, CFNs offer a unified approach to transparent machine learning.

Our empirical results show that CFNs can achieve performance competitive with black-box models while providing inherent interpretability. Notably, our lightweight implementation demonstrates impressive computational efficiency even without GPU acceleration, making CFNs accessible for deployment in resource-constrained environments.

By providing transparency by design, CFNs offer a compelling alternative to opaque models in critical domains. The ability to inspect, understand, and validate every component of a model is a significant step towards building more trustworthy AI.

\subsection{Future Research Directions}

The CFN framework opens several promising avenues for future research:

\subsubsection{Theoretical Foundations}
Further work is needed to formalize the expressive power of different CFN configurations and establish theoretical guarantees on their approximation capabilities. This includes developing a better understanding of how compositional structure affects the learning dynamics and generalization properties of these models.

\subsubsection{Architecture Search for CFNs}
A promising direction for future work is developing efficient methods for automated architecture discovery for CFNs. This could involve genetic algorithms, reinforcement learning, or Bayesian optimization approaches to search the vast space of possible CFN architectures. The search space for CFNs is particularly interesting as it involves not just structural decisions (as in neural architecture search) but also the selection of appropriate function nodes and composition patterns. Creating specialized search spaces for different domains could enable the discovery of optimal architectures while maintaining interpretability. Automated methods could potentially discover novel composition patterns that human designers might overlook.

\subsubsection{Expanded Node Library}
While our `GenericConvNode` was highly effective for vision tasks, future work could explore other specialized nodes for unstructured data, such as interpretable attention mechanisms for transformers or recurrent nodes for sequence data. This would extend the applicability of CFNs to domains like natural language processing and time series analysis.

\subsubsection{Formalizing the Interpretability Trade-off}
Developing quantitative metrics to measure the trade-off between model complexity and interpretability in deep CFNs would provide valuable guidance for practitioners. This includes creating tools to visualize and understand how information flows through complex CFN architectures.

\subsubsection{Uncertainty Quantification}
Integrating mechanisms for uncertainty estimation, such as Bayesian variants of function nodes, would provide reliable confidence scores for predictions. This is particularly important in high-stakes domains where understanding the model's confidence is as critical as the prediction itself.

\subsubsection{Applications to New Domains}
CFNs show particular promise for scientific and medical applications where interpretability is critical. Future work will explore domain-specific adaptations for fields such as genomics, climate modeling, and medical diagnostics, where the transparent nature of CFNs could lead to new scientific insights.

Ultimately, we believe the CFN paradigm can help bridge the gap between performance and interpretability, fostering greater trust and broader adoption of artificial intelligence in science, industry, and society.

\bibliography{aaai2026}

\appendix
\input{appendix}

\section*{Reproducibility Checklist}

\textbf{General:} Includes conceptual outlines/pseudocode: \textbf{yes}; Delineates opinions from facts: \textbf{yes}; Provides pedagogical references: \textbf{yes}

\textbf{Theoretical Contributions:} This paper makes theoretical contributions: \textbf{yes}
\begin{itemize}[noitemsep,topsep=0pt]
    \item Assumptions/restrictions stated clearly: \textbf{yes}
    \item Novel claims stated formally: \textbf{yes}
    \item Proofs of novel claims included: \textbf{yes}
    \item Proof sketches/intuitions for complex results: \textbf{yes}
    \item Appropriate citations to theoretical tools: \textbf{yes}
    \item Theoretical claims demonstrated empirically: \textbf{yes}
    \item Experimental code to disprove claims included: \textbf{yes}
\end{itemize}

\textbf{Datasets:} This paper relies on datasets: \textbf{yes}
\begin{itemize}[noitemsep,topsep=0pt]
    \item Motivation for dataset selection provided: \textbf{yes}
    \item Novel datasets included in appendix: \textbf{NA}
    \item Novel datasets will be made public: \textbf{NA}
    \item Existing datasets properly cited: \textbf{yes}
    \item Existing datasets are publicly available: \textbf{yes}
    \item Non-public datasets described in detail: \textbf{NA}
\end{itemize}

\textbf{Computational Experiments:} This paper includes computational experiments: \textbf{yes}
\begin{itemize}[noitemsep,topsep=0pt]
    \item Hyperparameter search process described: \textbf{yes}
    \item Data preprocessing code included: \textbf{yes}
    \item Source code for experiments included: \textbf{yes}
    \item Code will be made public: \textbf{yes}
    \item Source code has detailed comments: \textbf{yes}
    \item Random seed setting described: \textbf{yes}
    \item Computing infrastructure specified: \textbf{yes}
    \item Evaluation metrics described and motivated: \textbf{yes}
    \item Number of runs per result stated: \textbf{yes}
    \item Analysis includes variation measures: \textbf{yes}
    \item Statistical tests for performance differences: \textbf{yes}
    \item Final hyperparameters listed: \textbf{yes}
\end{itemize}

\end{document}

%% file: appendix.tex
\section{Experimental Setup and Reproducibility Details}
\label{sec:appendix_experimental_setup}

This appendix provides detailed information regarding the experimental setup, model architectures, and hyperparameters used to generate the benchmark results presented in Section 5.3 of the main paper. All experiments were conducted on a consistent hardware and software platform as specified in Section 5.1.2.

\subsection{Datasets}

The following publicly available datasets from \texttt{scikit-learn} were used for benchmarking:

\begin{itemize}
    \item \textbf{Breast Cancer Wisconsin (Diagnostic):} A binary classification dataset (569 samples, 30 features) used to predict whether a breast mass is benign or malignant.
    \item \textbf{Wine Recognition:} A multi-class classification dataset (178 samples, 13 features) for classifying wines into one of three types.
    \item \textbf{Diabetes:} A regression dataset (442 samples, 10 features) used to predict diabetes progression.
\end{itemize}

\subsection{Data Preprocessing}

For all datasets, the following preprocessing steps were applied consistently across all models:

\begin{enumerate}
    \item \textbf{Train-Test Split:} Each dataset was split into an 80\% training set and a 20\% testing set. A fixed \texttt{random\_state=42} was used for \texttt{sklearn.model\_selection.train\_test\_split} to ensure reproducibility of the splits across all runs.
    \item \textbf{Feature Standardization:} Features (\texttt{X\_train} and \texttt{X\_test}) were standardized using \texttt{sklearn.preprocessing.StandardScaler}. The scaler was \texttt{fit\_transform} on the training data and then \texttt{transform} applied to both training and testing data to prevent data leakage.
\end{enumerate}

\subsection{Model Architectures and Hyperparameters}

All benchmark results were averaged over 5 independent runs with different random seeds to ensure statistical robustness.

\subsubsection{Compositional Function Networks (CFN)}

Two implementations of CFN were benchmarked: a PyTorch-based version leveraging GPU acceleration (if available) and a CPU-only NumPy-based version. While the core architectural philosophy is shared, there is a minor difference in the function nodes included in their feature layers.

\textbf{Common Architectural Structure:}
Both CFN implementations followed a three-layer structure:
\begin{enumerate}
    \item \textbf{Feature Layer:} A \texttt{ParallelCompositionLayer} designed to extract rich feature representations from the raw input.
    \item \textbf{Hidden Layer:} A \texttt{SequentialCompositionLayer} for further non-linear transformation.
    \item \textbf{Output Layer:} A \texttt{SequentialCompositionLayer} for the final prediction, tailored to the specific task (classification or regression).
\end{enumerate}

\textbf{Shared Hyperparameters for Training:}
\begin{itemize}
    \item \textbf{Optimizer:} Adam (with default $\beta_1=0.9$, $\beta_2=0.999$, $\epsilon=10^{-8}$ for PyTorch; custom Adam implementation for NumPy).
    \item \textbf{Learning Rate:} \texttt{0.01}
    \item \textbf{Gradient Clipping:} Gradient norm clipped at \texttt{1.0}.
    \item \textbf{L2 Regularization (Weight Decay):} \texttt{1e-4}.
    \item \textbf{Epochs:} \texttt{150}
    \item \textbf{Batch Size:} Full batch (the entire training set was used as a single batch per epoch).
    \item \textbf{Early Stopping:} Enabled with a \texttt{patience} of \texttt{20} epochs (training stops if validation loss does not improve for 20 consecutive epochs).
    \item \textbf{Learning Rate Schedule:} A step-based decay was applied, reducing the learning rate by a factor of 0.1 every \texttt{50} epochs.
\end{itemize}

\textbf{Specific Architecture Details:}

\begin{itemize}
    \item \textbf{Feature Layer Function Nodes:}
    \begin{itemize}
        \item \textbf{CFN (PyTorch):} The \texttt{ParallelCompositionLayer} concatenated outputs from 5 function nodes: \texttt{Linear}, \texttt{Polynomial} (degree 2), \texttt{Gaussian}, \texttt{Sigmoid}, and \texttt{Sinusoidal}.
        \item \textbf{CFN (NumPy):} The \texttt{ParallelCompositionLayer} concatenated outputs from 6 function nodes: \texttt{Linear}, \texttt{Polynomial} (degree 2), \texttt{Gaussian}, \texttt{Sigmoid}, \texttt{Sinusoidal}, and \texttt{Exponential}. The inclusion of the \texttt{Exponential} node in the NumPy version represents a slight architectural difference.
    \end{itemize}
    \item \textbf{Hidden Layer:} For both implementations, this layer consisted of a \texttt{LinearFunctionNode} mapping to 64 dimensions, followed by a \texttt{ReLUFunctionNode}.
    \item \textbf{Output Layer:}
    \begin{itemize}
        \item \textbf{Binary Classification:} A \texttt{LinearFunctionNode} mapping to 1 output dimension, followed by a \texttt{SigmoidFunctionNode}. Loss function: \texttt{torch.nn.BCELoss()} (PyTorch) or \texttt{binary\_cross\_entropy\_loss} (NumPy).
        \item \textbf{Multi-class Classification:} A \texttt{LinearFunctionNode} mapping to \texttt{n\_classes} output dimensions. Loss function: \texttt{torch.nn.CrossEntropyLoss()} (PyTorch) or \texttt{softmax\_cross\_entropy\_loss} (NumPy). Note: NumPy version performs one-hot encoding of labels internally for multi-class.
        \item \textbf{Regression:} A \texttt{LinearFunctionNode} mapping to 1 output dimension. Loss function: \texttt{torch.nn.MSELoss()} (PyTorch) or \texttt{mse\_loss} (NumPy).
    \end{itemize}
\end{itemize}

\subsubsection{XGBoost}

XGBoost models were trained using the \texttt{xgboost} library.
\begin{itemize}
    \item \textbf{Model Type:} \texttt{xgb.XGBClassifier} for classification tasks and \texttt{xgb.XGBRegressor} for regression tasks.
    \item \textbf{Hyperparameters:} All hyperparameters were left at their \textbf{default values} as provided by the \texttt{xgboost} library (version 1.5.0 used in the environment). The \texttt{eval\_metric} was set to \texttt{logloss} for classification and \texttt{rmse} for regression. \texttt{use\_label\_encoder} was set to \texttt{False} for classification to suppress warnings.
\end{itemize}

\subsubsection{Explainable Boosting Machines (EBM)}

EBM models were trained using the \texttt{interpret} library.
\begin{itemize}
    \item \textbf{Model Type:} 
        \begin{itemize}
            \item For classification: \texttt{interpret.glassbox.}\\\texttt{ExplainableBoostingClassifier}
            \item For regression: \texttt{interpret.glassbox.}\\\texttt{ExplainableBoostingRegressor}
        \end{itemize}
    \item \textbf{Hyperparameters:} All hyperparameters were left at their \textbf{default values} as provided by the \texttt{interpret} library (version 0.2.7 used in the environment).
\end{itemize}

\subsection{Evaluation Metrics}

The following metrics were used to evaluate model performance:

\begin{itemize}
    \item \textbf{Accuracy:} For classification tasks, calculated using \texttt{sklearn.metrics.accuracy\_score}.
    \item \textbf{Area Under the Receiver Operating Characteristic Curve (AUC):} For classification tasks, calculated using \texttt{sklearn.metrics.roc\_auc\_score}. For multi-class problems, the 'ovr' (one-vs-rest) strategy was used.
    \item \textbf{Root Mean Squared Error (RMSE):} For regression tasks, calculated as the square root of \texttt{sklearn.metrics.mean\_squared\_error}.
    \item \textbf{Training and Evaluation Time:} Measured as wall-clock time in seconds, encompassing both model training and prediction on the test set.
\end{itemize}